\documentclass{article}
\usepackage{spconf,amsmath,graphicx}
\usepackage{booktabs} 
\usepackage{pgfplots}
\usepackage{multirow}
\pgfplotsset{compat=1.17}
\usepackage{float}

\usepackage{enumitem}
\setlist{nosep, leftmargin=14pt}

\usepackage{mwe}

\title{CXR-LT 2026 Challenge: Projection-Aware Multi-Label and Zero-Shot Chest X-Ray Classification}

\name{
Juno Cho$^{1\dagger}$,
Dohui Kim$^{2\dagger}$,
Mingeon Kim$^{1\dagger}$,
Hyunseo Jang$^{3\dagger}$,
Chang Sun Lee$^{4}$,
Jong Chul Ye$^{4\ast}$
}

\address{
$^{1}$ KAIST, Daejeon, South Korea \\
$^{2}$ GIST, Gwangju, South Korea \\
$^{3}$ Korea University, Seoul, South Korea \\
$^{4}$ KAIST Graduate School of AI (GSAI), Seoul, South Korea
}

\begin{document}
\maketitle
\begingroup
\renewcommand\thefootnote{}
\footnotetext{\footnotesize $^{\dagger}$ Equal contribution. $^{\ast}$ Corresponding author. Work conducted during the KAIRI internship at KAIST. Dohui Kim graduated from GIST (Feb. 2026). \\ Accepted to the IEEE International Symposium on Biomedical Imaging (ISBI) 2026 CXR-LT Challenge. \copyright 2026 IEEE. Personal use of this material is permitted. Permission from IEEE must be obtained for all other uses, in any current or future media, including reprinting/republishing this material for advertising or promotional purposes, creating new collective works, for resale or redistribution to servers or lists, or reuse of any copyrighted component of this work in other works.}
\endgroup
\begin{abstract}
This challenge tackles multi-label classification for known chest X-ray (CXR) lesions and zero-shot classification for unseen ones. To handle diverse CXR projections, we integrate projection-specific models via a classification network into a unified framework. For zero-shot classification (Task 2), we extend CheXzero with a novel dual-branch architecture that combines contrastive learning, Asymmetric Loss (ASL), and LLM-generated descriptive prompts. This effectively mitigates severe long-tail imbalances and maximizes zero-shot generalization. Additionally, strong data and test-time augmentations (TTA) ensure robustness across both tasks.
\end{abstract}

\begin{keywords}
Projection-Aware, Zero-Shot Classification, CLIP, Asymmetric Loss, Descriptive Prompt, Ensemble
\end{keywords}

\section{Introduction}
\label{sec:intro}
The CXR-LT 2026 Challenge~\cite{CXRLT2026} addresses severe long-tailed data distributions and annotation scarcity in real-world CXR diagnosis through two interconnected tasks. \textbf{Task 1} targets multi-label classification for 30 known lesions, where we mitigate class imbalances using a projection-aware routing network paired with robust ensembling. For \textbf{Task 2}, which demands zero-shot classification for six entirely unseen diseases without explicit supervision, our contributions are threefold: (1) a dual-branch framework integrating global semantic alignment with ASL for fine-grained discrimination; (2) medical concept enrichment via LLM-based prompt ensembling; and (3) a strict proxy validation strategy to guarantee leak-free out-of-distribution (OOD) generalization.

\section{Related Works}
\label{sec:relwork}

Deep learning for CXR classification has evolved from convolutional networks to transformer-based models. CNNs such as DenseNet~\cite{DenseNet}, EfficientNet~\cite{EfficientNet}, and ConvNeXt~\cite{ConvNeXt,ConvNeXtV2} remain widely used for multi-label tasks due to strong inductive biases. Transformer variants including ViT~\cite{ViT}, CaiT~\cite{CaiT}, and Swin Transformer~\cite{Swin,SwinV2} capture long-range dependencies via self-attention and perform well on high-resolution medical images.

In CXR analysis, CheXNet~\cite{CheXNet} demonstrated large-scale multi-label disease detection on ChestX-ray14, while CheXpert~\cite{CheXpert} addressed label uncertainty. Subsequent works mitigate class imbalance and weak supervision in long-tailed datasets using attention mechanisms or region-aware pooling such as PCAM~\cite{PCAM,COVIDViT}.

Zero-shot classification enables recognition of unseen diseases without additional annotations. CLIP~\cite{CLIP} and its medical adaptation CheXzero~\cite{CheXzero} align images and text in a shared embedding space for zero-shot inference. However, contrastive learning is sensitive to severe class imbalance in medical data. Asymmetric Loss (ASL)~\cite{ASL} mitigates this issue, while recent studies leverage Large Language Models (LLMs) to generate descriptive prompts \cite{platypus, BiomedCoOp} for improved zero-shot generalization under domain shifts.

\section{Methods}
\label{sec:meth}

\subsection{Task 1: Multi-label Classification}
\label{ssec:meth-task1}

\begin{figure*}[t]
    \centering
    \includegraphics[width=0.80\linewidth]{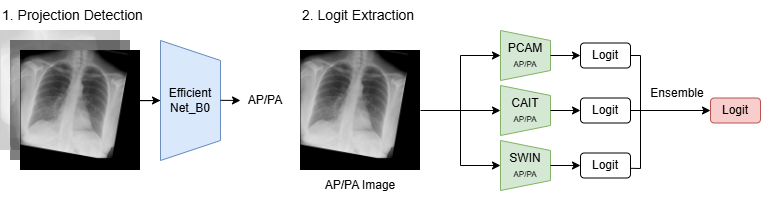}
    \caption{Multi-label classification diagram.}
    \label{fig:diagram}
\end{figure*}
Training data were split by projection into  AP/PA and Lateral (L) subsets, specializing models on each view. An EfficientNet Router predicts the projection and forwards the image to the corresponding ensemble to produce 30-dimensional logits (Fig.~\ref{fig:diagram}). 
Each projection branch is an ensemble of three architectures:
\begin{itemize}
     \item \textbf{PCAM}: ConvNeXt-v2 serves as a CNN encoder followed by PCAM pooling \cite{PCAM}, which generates label-wise heatmaps to guide attention. Using only a 1×1 convolution encourages discriminative representations.

     \item \textbf{CAIT}: ConvNeXt-v2 is first pre-trained with the PCAM framework and then frozen. CaiT is applied to the projected CNN features, leveraging transformer modeling on semantically summarized feature vectors as motivated by \cite{COVIDViT}. The final CLS token is mapped to logits via a linear layer.

     \item \textbf{SWIN}: Swin Transformer employs hierarchical self-attention within shifted windows, reducing the quadratic cost of global attention. Its locality-preserving design and scalability to high-resolution inputs make it suitable for CXR lesion detection.

Data were split 9:1 for training and validation. Strong augmentations (horizontal flip, rotation, translation, scaling) were applied during training. Model selection was based on mAP, mAUC, and BCE loss on the augmented validation set. TTA (horizontal flip) was used to average logits, and final ensemble coefficients were determined via grid search at the logit level.
\end{itemize}

\subsection{Task 2: Zero-Shot Classification}

\subsubsection{Dual-Branch Hybrid Architecture}

\begin{figure}[h]
    \centering
    \includegraphics[width=1.0\linewidth]{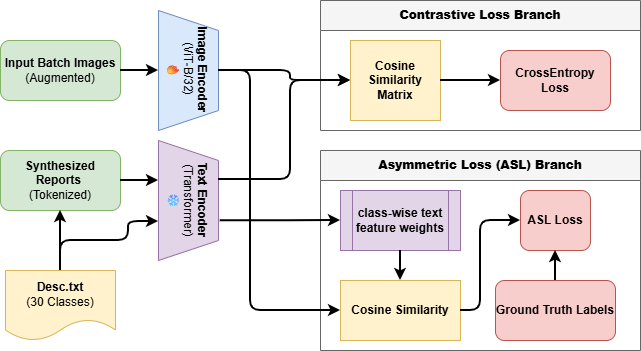}
    \caption{Dual-Branch Hybrid Architecture.}
    \label{fig:dual_way_arch}
\end{figure}

We propose a dual-branch architecture (Fig. \ref{fig:dual_way_arch}) consisting of a Contrastive Loss branch and an Asymmetric Loss branch:
\begin{itemize}
\item \textbf{Contrastive Loss Branch:} Learns global alignment between images and label-synthesized reports, providing a representation baseline for zero-shot recognition.

\item \textbf{Asymmetric Loss (ASL) Branch:} Introduced to mitigate the sensitivity of contrastive learning to severe data imbalance. The total loss is defined as
\begin{equation}
\mathcal{L}_{total} = \mathcal{L}_{con} + \alpha \mathcal{L}_{ASL}
\label{eq:total_loss}
\end{equation}
\begin{equation}
\mathcal{L}_{ASL} = \frac{1}{K} \sum_{k=1}^{K} \begin{cases} 
-(1 - p_k)^{\gamma_{+}} \log(p_k) & \text{if } y_k=1 \\
-(p_k)^{\gamma_{-}} \log(1 - p_k) & \text{if } y_k=0
\end{cases}
\end{equation}
where $\alpha$ balances the two losses, $K=30$, $y_k \in \{0,1\}$ is the ground-truth label and $p_k$ the predicted probability for class $k$. $\gamma_{-}$ and $\gamma_{+}$ are focusing parameters; setting $\gamma_{-} > \gamma_{+}$ suppresses easy negatives and emphasizes hard positives, improving sensitivity for rare classes.
\end{itemize}

\subsubsection{Description-Based Prompting \& OOD Inference Strategy}

We used LLM-generated clinical descriptions instead of simple class names. This encouraged the text encoder to capture fine-grained pathology semantics (e.g., “diffuse decreased bone density” for Osteopenia) and strengthened cross-modal alignment.

\begin{itemize}
\item \textbf{Training Phase -- Dynamic Shuffling:}
For multi-label samples, descriptions were concatenated and randomly shuffled to prevent lexical overfitting and promote semantic understanding.

\item \textbf{Inference Phase -- Hybrid Prompting for OOD Generalization:}
At inference, we used both class names and descriptive prompts. The description-based training induces a semantically structured embedding space, enabling alignment with unseen OOD prompts without explicit supervision. Prompt ensembling and TTA were further applied for robustness.

\end{itemize}

\section{Experiments}
\label{sec:expdet}
\subsection{Task 1: Multi-Label Classification}
\label{ssec:expdet-task1}
To minimize distortion, images were zero-padded and resized to $512 \times 512$ (uint16) preserving aspect ratios. We applied random augmentations including horizontal flip ($p=0.5$), rotation ($\pm 20^\circ$), scaling ($[0.9, 1.1]$), translation ($\le 5\%$), and CLAHE ($p=0.3$). The Swinv2-tiny and ConvNeXt-v2-tiny backbones were initialized with ImageNet pre-trained weights, using standard normalization for 3-channel inputs.

\subsection{Task 2: Zero-Shot Classification}
For Task~2, both the image and text encoders were initialized with CheXzero~\cite{CheXzero} weights and fine-tuned on in-distribution CXRs (resized to $224 \times 224$). The model was optimized using AdamW (learning rate $10^{-6}$, weight decay 0.01), a cosine annealing scheduler ($T_{max}=7$), and EMA (decay=0.999). Lacking target OOD labels, we adopted a leak-free 3-fold proxy validation strategy. We designated 6 of the 30 labels as proxy unseen classes, removing their samples during training to prevent data leakage. The model, trained on the remaining 24 classes, was evaluated via AP on three held-out proxy groups: Group A (Structural Anomalies \& Devices) for fine-grained deformations, Group B (High Prevalence Diseases) for baseline stability, and Group C (Critical \& Rare Conditions) to assess resilience against catastrophic forgetting in long-tail distributions.

\section{Results}
\label{sec:results}

\subsection{Task 1: Multi-Label Classification}
\label{ssec:results-task1}

Table \ref{tab:results_task1} summarizes the performance of projection-aware and mixed-projection models. Ensemble weights were determined by grid search using validation metrics. All L models showed lower performance than their AP/PA counterparts.

The EfficientNet-B0 router achieved 98.65\% validation accuracy, and the integrated routing framework obtained an mAP of 0.39456 on the entire validation set. The optimal ensemble weights were PCAM : SWIN : CAIT $=0.40:0.40:0.20$ for AP/PA and $0.45:0.45:0.10$ for L.

Despite routing and ensembling, the framework is computationally efficient, requiring $\sim$77 ms/image in inference.

In the router ablation study, models trained on mixed projections showed notable performance improvements. Table \ref{tab:results_task1} reports the results evaluated separately on each projection. Although L projections remained more challenging than AP/PA, consistent gains in both projections suggest complementary learning across projections.

Also, we evaluated cross-dataset generalization on the external MIMIC dataset \cite{mimic_physionet} with 7 overlapping labels. The projection-aware model achieved mAP of 0.1960 (AP/PA) and 0.1243 (L), while the mixed model achieved 0.2063 and 0.1210, respectively.

Although the mixed model slightly outperformed on AP/PA, the projection-aware framework showed better robustness on Lateral projections. We hypothesize that mixed training, while leveraging shared features, is biased toward the dominant AP/PA distribution, degrading performance on the minority L view. In contrast, the routing framework preserves view-specific features by isolating Lateral representations. This suggests that projection-aware routing better maintains robustness across projections, supporting its effectiveness for generalization to unseen test distributions.

\begin{table}[ht]
\centering
\caption{Validation results of projection-aware and mixed-projection models.}
\label{tab:results_task1}
\resizebox{\columnwidth}{!}{
\begin{tabular}{lllcccc}
\toprule
\textbf{Setting} & \textbf{Proj} & \textbf{Model} & \textbf{mAP} & \textbf{BCE} & \textbf{mAUC} & \textbf{mECE} \\
\midrule
\multirow{8}{*}{Projection aware}
 & \multirow{4}{*}{AP/PA} & PCAM & 0.4011 & 0.0871 & 0.9117 & 0.0067 \\
 &                        & SWIN & 0.4028 & 0.0866 & 0.9104 & 0.0052 \\
 &                        & CAIT & 0.3925 & 0.0868 & 0.9051 & 0.0067 \\
 &                        & Ensemble & \textbf{0.4185} & \textbf{0.0841} & \textbf{0.9182} & \textbf{0.0045} \\
 \cmidrule(lr){2-7}
 & \multirow{4}{*}{L}     & PCAM & 0.3336 & 0.1061 & 0.8681 & 0.0111 \\
 &                        & SWIN & 0.3045 & 0.1051 & 0.8621 & 0.0099 \\
 &                        & CAIT & 0.1601 & 0.1318 & 0.7591 & 0.0101 \\
 &                        & Ensemble & \textbf{0.3586} & \textbf{0.1028} & \textbf{0.8753} & \textbf{0.0087} \\
\midrule
\multirow{8}{*}{Mixed}
 & \multirow{4}{*}{AP/PA} & PCAM & 0.4283 & 0.0866 & 0.9092 & 0.0069 \\
 &                        & SWIN & 0.3954 & 0.0862 & 0.9111 & 0.0073 \\
 &                        & CAIT & 0.3945 & 0.0872 & 0.9070 & 0.0058 \\
 &                        & Ensemble & \textbf{0.4433} & \textbf{0.0835} & \textbf{0.9161} & \textbf{0.0058} \\
 \cmidrule(lr){2-7}
 & \multirow{4}{*}{L}     & PCAM & 0.3642 & 0.1027 & 0.8553 & 0.0089 \\
 &                        & SWIN & 0.3654 & 0.1018 & 0.8569 & 0.0105 \\
 &                        & CAIT & 0.3592 & 0.1045 & 0.8518 & 0.0075 \\
 &                        & Ensemble & \textbf{0.3985} & \textbf{0.0989} & \textbf{0.8650} & \textbf{0.0054} \\
\bottomrule
\end{tabular}
}
\end{table}

\subsection{Task 2: Zero-Shot Classification}

\begin{figure}[ht]
    \centering
    \begin{tikzpicture}
    \begin{axis}[
        width=0.8\columnwidth, height=4.5cm,
        scale only axis,
        xlabel={Training Epochs},
        ylabel={mAP Score},
        xmin=1, xmax=7,
        ymin=0.26, ymax=0.30,
        xtick={1,2,3,4,5,6,7},
        ymajorgrids=true,
        grid style=dashed,
        tick label style={font=\scriptsize},
        label style={font=\small},
        legend style={at={(0.95,0.95)}, anchor=north east, font=\scriptsize},
        mark size=1.5pt ]

        \addplot[color=black, dashed, very thick, mark=none]
        coordinates {(1,0.2635)(7,0.2635)}; 
        \addlegendentry{Baseline}

        \addplot[color=blue, mark=square*, thick]
        coordinates {
            (1,0.2695)(2,0.2707)(3,0.2691)(4,0.2667)(5,0.2648)(6,0.2630)(7,0.2615)
        };
        \addlegendentry{$\alpha=0$}

        \addplot[color=green!60!black, mark=triangle*, thick]
        coordinates {
            (1,0.2763)(2,0.2855)(3,0.2832)(4,0.2771)(5,0.2724)(6,0.2684)(7,0.2649)
        };
        \addlegendentry{$\alpha=1.0$}

        \addplot[color=red, mark=*, thick]
        coordinates {
            (1,0.2775)(2,0.2882)(3,0.2865)(4,0.2816)(5,0.2783)(6,0.2747)(7,0.2710)
        };
        \addlegendentry{$\alpha=1.5$}

        \end{axis}
    \end{tikzpicture}
    \caption{\textbf{Validation mAP on Proxy Group A.} ASL ($\alpha=1.5$) consistently outperforms $\alpha=0$ and the baseline.}
    \label{fig:validation_curve}
\end{figure}
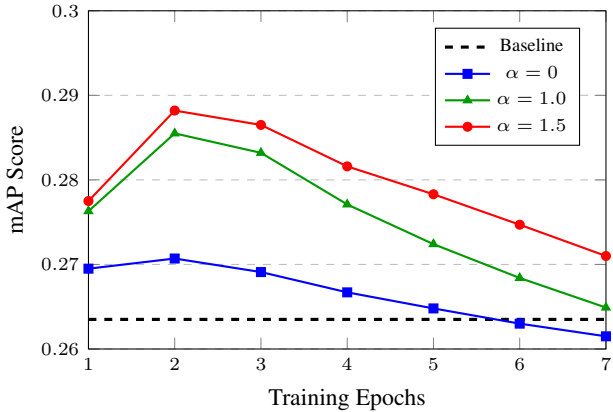

Table \ref{tab:zero_shot_results} shows the impact of label quality and ASL on OOD performance. Fine-tuning with either all generated labels or strictly physician-verified labels yielded identical overall performance increases. Building upon this filtered dataset, we evaluated the proposed Dual-Branch architecture with ASL. Setting $\alpha=1.5$ achieved the best overall average mAP of 0.3391, demonstrating consistent improvements across the proxy groups, particularly in structural anomaly detection. Figure \ref{fig:validation_curve} illustrates the validation trajectory on Group A. While standard fine-tuning ($\alpha=0$) shows limited gain, ASL ($\alpha=1, 1.5$) significantly boosts performance.

To evaluate calibration, we analyzed the Expected Calibration Error (ECE) at the checkpoints yielding the highest zero-shot mAP for each proxy set. Results show that ASL ($\alpha=1.5$) effectively enhances reliability, reducing the average ECE across the three proxy groups from 0.2902 (baseline, $\alpha=0$) to 0.2417.
Regarding computational and resource efficiency, our framework is highly feasible for clinical use. The ViT-B/32 backbone maintains a lightweight footprint, and the ASL branch adds zero inference overhead as it is utilized exclusively during training. Baseline inference takes $\sim$17 ms per image on a single NVIDIA RTX 2080 Ti GPU. Pre-computing multiple LLM prompt embeddings for the prompt ensemble adds negligible delay, totaling $\sim$18 ms. Even fully deployed with TTA, latency is only 27 ms, ensuring rapid clinical diagnosis.

\begin{table}[ht]
    \centering
    \caption{Impact of data quality and Asymmetric Loss (ASL) on zero-shot mAP. (FT: Standard Fine-Tuning, $\alpha=0$)}
    \label{tab:zero_shot_results}
    \resizebox{\columnwidth}{!}{%
    \begin{tabular}{lcccc}
    \toprule
    \textbf{Method} & \textbf{Group A} & \textbf{Group B} & \textbf{Group C} & \textbf{Avg.} \\
    \midrule
    Baseline (Zero-shot) & 0.2635 & 0.3790 & 0.3134 & 0.3186 \\
    FT (All Labels) & 0.2680 & 0.3852 & 0.3126 & 0.3219 \\
    FT (Physician Only) & 0.2707 & 0.3852 & 0.3098 & 0.3219 \\
    \textbf{Dual-Branch ($\alpha=1.5$)} & \textbf{0.2882} & \textbf{0.4139} & \textbf{0.3151} & \textbf{0.3391} \\
    \bottomrule
    \end{tabular}
    }
\end{table}

\section{Conclusion}
\label{sec:conclusion}
We proposed a robust framework for long-tailed and zero-shot CXR classification. In Task 1, projection-aware routing and ensemble strategy effectively mitigated class imbalances, achieving test mAP of 0.4827, mAUC of 0.9186 and F1 of 0.3162, demonstrating the potential for further improvement through mixed-projection modeling. In Task 2, the proposed dual-branch architecture, combining Asymmetric Loss (ASL) with LLM-generated descriptive prompts, demonstrated strong zero-shot generalization with test mAP of 0.3106, AUC of 0.6705, and F1 of 0.2025. Evaluated on the official hidden test set, these results took 2nd place in both tasks at the ISBI 2026 CXR-LT Challenge.

\vspace{2mm} 
\noindent \textbf{Compliance with Ethical Standards:} This study retrospectively used de-identified data from the CXR-LT Challenge. Access was granted per the Terms of Use; ethical approval was waived as the public dataset lacks identifiable personal information.

\noindent \textbf{Acknowledgments:} 
 
This research was supported by the Institute of Information \& Communications Technology Planning \& Evaluation (IITP) grant funded by the Korea government (MSIT) (RS-2025-02304967, AI Star Fellowship (KAIST)). 
The authors declare no conflicts of interest.

\bibliographystyle{IEEEbib}
\bibliography{strings,refs}

\end{document}